\documentclass[runningheads,a4paper]{llncs}

\usepackage{amssymb}
\usepackage{graphicx}

\usepackage{amsfonts,amsmath}
\usepackage{tikz}
\usepackage{algorithm}
\usepackage{algorithmic}
\usepackage{color}
\usepackage[width=122mm,left=12mm,paperwidth=146mm,height=193mm,top=12mm,paperheight=217mm]{geometry}
\usepackage{times}
\usepackage{epsfig}
\usepackage{enumitem}
\usepackage{array}
\usepackage{float}
\usepackage{hyperref}

\newcolumntype{C}[1]{>{\centering\arraybackslash}p{#1}}

\usepackage{url}
 
\newcommand{\keywords}[1]{\par\addvspace\baselineskip
\noindent\keywordname\enspace\ignorespaces#1}

\begin{document}

\mainmatter  

\title{Ranking to Learn: \\\large Feature Ranking and Selection via Eigenvector Centrality}

\titlerunning{Ranking to Learn: Feature Ranking and Selection via Eigenvector Centrality}
\authorrunning{Roffo \& Melzi}
\author{\qquad Giorgio Roffo$^{1,2}$ \qquad \qquad  \qquad \quad Simone Melzi$^2$\\
Giorgio.Roffo@glasgow.ac.uk \qquad Simone.Melzi@univr.it }


\institute{School of Computing Science, University of Glasgow, United Kingdom
\and
Department of Computer Science, University of Verona, Italy}
\renewcommand{\algorithmicrequire}{\textbf{Input:}}
\renewcommand{\algorithmicensure}{\textbf{Output:}}
%
%

\toctitle{Ranking to Learn: Feature Ranking and Selection via Eigenvector Centrality}
\tocauthor{Giorgio Roffo}
\maketitle

\begin{abstract}
In an era where accumulating data is easy and storing it inexpensive, feature selection plays a central role in helping to reduce the high-dimensionality of huge amounts of otherwise meaningless data. In this paper, we propose a graph-based method for feature selection that ranks features by identifying the most important ones into arbitrary set of cues. Mapping the problem on an affinity graph - where features are the nodes - the solution is given by assessing the importance of nodes through some indicators of centrality, in particular, the \emph{Eigenvector Centrality (EC)}. The gist of EC is to estimate the importance of a feature as a function of the importance of its neighbors. Ranking central nodes individuates candidate features, which turn out to be effective from a classification point of view, as proved by a thoroughly experimental section. Our approach has been tested on 7 diverse datasets from recent literature (e.g., biological data and object recognition, among others), and compared against filter, embedded and wrappers methods. The results are remarkable in terms of accuracy, stability and low execution time.
\keywords{Feature Selection, Ranking, High Dimensionality, Data Mining}
\end{abstract}

\section{Introduction}\vspace{-0.3cm}

As data collection technologies advance and computer power grows, a torrent of data is generated in almost every field computers are used~\cite{BolonCanedo201533}. Because the volume, velocity, variety and complexity of datasets is continuously increasing, pattern recognition methodologies have become indispensable in order to extract useful information from huge amounts of otherwise meaningless data.

\begin{figure*}[t]
\centering
\includegraphics[width=1.0\linewidth]{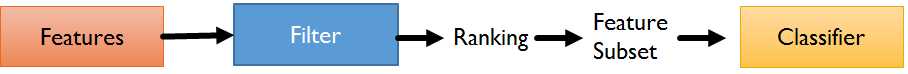}
\caption{Filter Methods: the selection of features is independent of the classifier used. They rely on the general characteristics of the training data to select features with independence of any predictor.}\label{fig:families1}
\end{figure*}

Feature Selection (FS) is one of the long existing methods that deals with these problems~\cite{guyon2006feature}. Its objective is to select a minimal subset of those attributes that allows a problem to be clearly defined. By choosing a minimal subset of features, irrelevant and redundant features are removed according to some reasonable criteria so that the original task can be achieved equally well, if not better. FS techniques can be partitioned into three classes~\cite{guyon2006feature}: \emph{wrappers} (see Fig.~\ref{fig:families2}), which use classifiers to score a given subset of features; \emph{embedded} methods  (in Fig.~\ref{fig:families3}), which inject the selection process into the learning of the classifier; and \emph{filter} methods (see Fig.~\ref{fig:families1}), which analyze intrinsic properties of data, ignoring the classifier. Filters are also (relatively) robust against overfitting. 

Most of these methods can perform two operations, \emph{ranking} and \emph{subset selection}: in the former, the importance of each individual feature is evaluated, usually by neglecting potential interactions among the elements of the joint set~\cite{duch2004comparison}; in the latter, the final subset of features to be selected is provided. In some cases, these two operations are performed sequentially (first the ranking, then the selection) ~\cite{Bradley98featureselection,Grinblat:2010,Guyon:2002,liu2008,Hutter:02feature};  in other cases, only the selection is carried out \cite{Quanquanjournals}. Usually, the subset selection is supervised, while in the ranking case, methods can be supervised or not. FS is \emph{NP-hard}~\cite{guyon2006feature}; if there are $n$ features in total, the goal is to select the optimal subset of $m\!\!\ll \!\!n$, by evaluating $\binom{n}{m}$ combinations; therefore, suboptimal search strategies are considered (see Section \ref{sec:SoA} for an overview). With the filters, features are first considered individually, ranked, and then a subset is extracted, some examples are Mutual Information~\cite{Hutter:02feature}, Relief-F~\cite{liu2008}, Inf-FS~\cite{Roffo_2015_ICCV,RoffoBMVC2016} unsupervised and not ~\cite{Obertino7552347}, and mRMR~\cite{Peng05featureselection}. Conversely, with wrapper and embedded methods, subsets of features are sampled, evaluated, and finally kept as the final output, for instance, FSV ~\cite{Bradley98featureselection,Grinblat:2010} and SVM-RFE~\cite{Guyon:2002}. 

In this work, we propose a novel graph-based feature selection algorithm that ranks features according to a graph centrality measure (Eigenvector centrality~\cite{Bonacich}). The main idea behind the method is to map the problem on an affinity graph, and to model pairwise relationships among feature distributions by weighting the edges connecting them. 

The novelty of the proposed method in terms of the state of the art is that it assigns a score of ``importance" to each feature by taking into account all the other features mapped as nodes on the graph, bypassing the combinatorial problem in a methodologically sound fashion. Indeed, eigenvector centrality differs from other measurements (e.g., degree centrality) since a node - feature - receiving many links does not necessarily have a high eigenvector centrality. The reason is that not all nodes are equivalent, some are more relevant than others, and, reasonably, endorsements from important nodes count more (see Section~\ref{sec:varphi} ). Noteworthy, another important contribution of this work is the scalability of the method. Indeed, centrality measurements can be implemented using the Map Reduce paradigm ~\cite{Kang_centralitiesin,Lerman:2010,MapReduceBetweenness}, which makes the algorithm prone to a possible distributed version~\cite{Rawat2015}.

Our approach is extensively tested on 7 benchmarks of cancer classification and prediction on genetic data (\emph{Colon}~\cite{alon}, \emph{Prostate}~\cite{Golub99}, \emph{Leukemia}~\cite{Golub99},\emph{Lymphoma}~\cite{Golub99}), handwritten recognition (GINA~\cite{GINA}), generic feature selection datasets (MADELON~\cite{guyon2004result}), and object recognition (PASCAL VOC 2007~\cite{pascal-voc-2007}). We compare the proposed method on these data, while comparing it against seven efficient approaches under different conditions (number of features selected and number of training samples considered), overcoming all of them in terms of ranking stability or classification accuracy.

Finally, we provide an open and portable library of feature selection algorithms, integrating the methods with uniform input and output formats to facilitate large scale performance evaluation. The \textit{Feature Selection Library} (FSLib Matlab Toolbox \footnote{The FSLib is publicly available on File Exchange - MATLAB Central at:\\ \url{https://it.mathworks.com/matlabcentral/fileexchange/56937-feature-selection-library}}) and interfaces are fully documented. The library integrates directly with MATLAB, a popular language for machine learning and pattern recognition research.

The rest of the paper is organized as follows. A brief overview of the related literature is given in Section~\ref{sec:SoA}, mostly focusing on the comparative approaches we consider in this work. Our feature selection algorithm is described in Section~\ref{sec:method}. Graph construction and weighting are presented in Section~\ref{sec:Graph} and Section~\ref{sec:varphi} respectively, while the employed Eigenvector centrality is discussed in Section~\ref{sec:EV}. Section~\ref{sec:exp} contains the experimental evaluations and results. Finally, conclusions are provided in Section~\ref{sec:conc}.

\section{Related Work}\label{sec:SoA}
\begin{figure*}[t]
\centering
\includegraphics[width=1.0\linewidth]{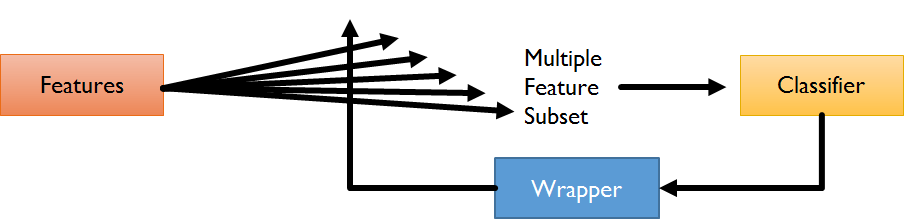}
\caption{Wrapper models involve optimizing a predictor as part of the selection process. They tend to give better results but filter methods are usually computationally less expensive than wrappers.}\label{fig:families2}
\end{figure*}

\vspace{-0.05cm}
Since the mid-1990s, few domains explored used more than 50 features.  The situation has changed considerably in the past few years and most papers explore domains with hundreds to tens of thousands of features. New approaches are proposed to address these challenging tasks involving many irrelevant and redundant cues and often comparably few training examples. Among the most used FS strategies, \emph{Relief-F}~\cite{liu2008} is an iterative, randomized, and supervised approach that estimates the quality of the features according to how well their values differentiate data samples that are near to each other; it does not discriminate among redundant features (i.e., may fail to select the most useful features), and performance decreases with few data. Similar problems affect SVM-RFE (RFE)~\cite{Guyon:2002}, which is a wrapper method (see Fig.~\ref{fig:families2}) that
selects features in a sequential, backward elimination manner, ranking high a feature if it strongly separates the samples by means of a linear SVM.

Batti \cite{Battiti1994} has developed the Mutual Information-Based Feature Selection (MIFS) criterion, where the features are selected in a greedy manner. Given a set of existing selected features, at each step it locates the feature $x_i$ that maximizes the relevance to the class. The selection is regulated by a proportional term $\beta$ that measures the overlap information between the candidate feature and existing features.  In \cite{Zhang2011} the authors proposed a \textit{graph-based} filter approach to feature selection, that constructs a graph in which each node corresponds to each feature, and each edge has a weight corresponding to mutual information (MI) between features connected by that edge. This method performs dominant set clustering to select a highly coherent set of features and then it selects features based on the multidimensional interaction information (MII). Another effective yet fast filter method is the \textit{Fisher} method~\cite{Quanquanjournals}, it computes a score for a feature as the ratio of interclass separation and intraclass variance, where features are evaluated independently, and the final feature selection occurs by aggregating the $m$ top ranked ones.  Other widely used filters are based on mutual information, dubbed \emph{MI} here \cite{Hutter:02feature}, which considers as a selection criterion the mutual information between the distribution of the values of a given feature and the membership to a particular class. Mutual information provides a principled way of measuring the mutual dependence of two variables, and has been used by a number of researchers to develop information theoretic feature selection criteria. Even in the last case, features are evaluated independently, and the final feature selection occurs by aggregating the $m$ top ranked ones. Maximum-Relevance Minimum-Redundancy criterion (MRMR) \cite{Peng05featureselection} is an efficient incremental search algorithm. Relevance scores are assigned by maximizing the joint mutual information between the class variables and the subset of selected features. The computation of the information between high-dimensional vectors is impractical, as the time required becomes prohibitive. To face this problem the mRMR propose to estimate the mutual information for continuous variables using Parzen Gaussian windows. This estimate is based on a heuristic framework to minimize redundancy and uses a series of intuitive measures of relevance and redundancy to select features. Note, it is equivalent to MIFS with $\beta = \frac{1}{n - 1}$, where $n$ is the number of features.
Selecting features in unsupervised learning scenarios is a much harder problem, due to the absence of class labels that would guide the search for relevant information. In this scenario, we compare our approach against the recent unsupervised graph-based filter dubbed Inf-FS~\cite{Roffo_2015_ICCV}. In the Inf-FS formulation, each feature is a node in the graph, a path is a selection of features, and the higher the centrality score, the most important (or most different) the feature. It assigns a score of ``importance" to each feature by taking into account all the possible feature subsets as paths on a graph. Another unsupervised method is the Laplacian Score (LS)~\cite{HCN05a}, where the importance of a feature is evaluated by its power of locality preserving. In order to model the local geometric structure, this method constructs a nearest neighbor graph. LS algorithm seeks those
features that respect this graph structure.
 \begin{figure*}[t]
\centering
\includegraphics[width=1.0\linewidth]{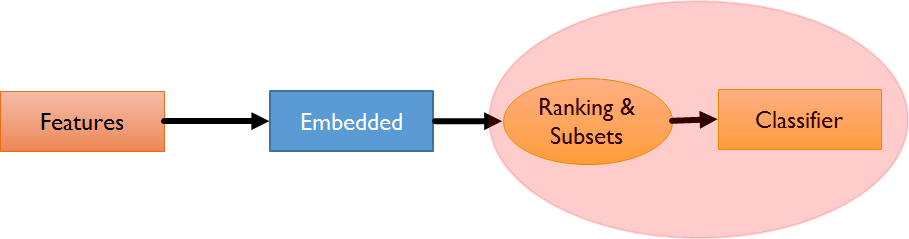}
\caption{In embedded methods the learning part and the feature selection part can not be separated.}\label{fig:families3}
\end{figure*} 
Finally, for the embedded method (see Fig.~\ref{fig:families3}), we include the \emph{feature selection via concave minimization} (\emph{FSV})~\cite{Bradley98featureselection}, where the selection process is injected \emph{into} the training of an SVM by a linear programming technique.

\section{Proposed Method}\label{sec:method}

\subsection{Building the Graph}\label{sec:Graph}
Given a set of features $X = \{  x^{(1)}, ..., x^{(n)} \}$ we build an undirected graph $G = (V, E)$; where $V$ is the set of vertices corresponding, one by one, to each variable $x$. $E$ codifies (weighted) edges among features. Let the adjacency matrix $A$ associated with $G$ define the nature of the weighted edges: each element $a_{ij}$ of $A$, $1\leq i,j \leq n$, represents a pairwise potential term. Potentials can be represented as a binary function $\varphi(x^{(i)},x^{(j)})$ of the nodes $x^{(k)}$ such as:

\begin{equation}\label{eq:partzero}
  a_{ij} = 	\varphi(x^{(i)},x^{(j)}).
\end{equation}

The graph can be weighted according to different heuristics, therefore the function $\varphi$ can be handcrafted or automatically learned from data.

\subsection{$\varphi$-Design}\label{sec:varphi}

The design of the $\varphi$ function is a crucial operation. In this work, we weight the graph according to good reasonable criteria, related to class separation, so as to address the classification problem. In other words, we want to rank features according to how well they discriminate between two classes. Hence, we draw upon best-practice in FS and propose an ensemble of two different measures capturing both relevance (supervised) and redundancy (unsupervised) proposing a kernelized-based adjacency matrix. Before continuing with the discussion, note that each feature distribution $x^{(i)}$ is normalized so as to sum to $1$. 

Firstly, we apply the Fisher criterion:
 \[
	 f_i = \frac{\left | \mu_{i,1} - \mu_{i,2} \right |^2}{ \sigma_{i,1}^2+\sigma_{i,2}^2},
\]
where $\mu_{i,\textit{C}}$ and $\sigma_{i,\textit{C}}$ are the mean and standard deviation, respectively, assumed by the $i$-th feature when considering the samples of the $\textit{C}$-th class. The higher $f_i$, the more discriminative the $i$-th feature. However, a natural generalization of this score into a multi-class framework is given by
 \[
	f_i = \frac{ \sum_{c=1}^\textbf{C}  (\mu_{i,c} - \mu_{i})^2 }{\sigma_{i}^2}
\]
where $\mu_{i}$ and $\sigma_{i}$ denote the mean and standard deviation of the whole data set corresponding to the $i$-th feature (i.e.,  $\sigma_{i}^2 = \sum_{c=1}^\textbf{C}  (\sigma_{i,c})^2$).  

Because we are given class labels, it is natural that we want to keep only the features that are related to or lead to these classes. Therefore, we use mutual information to obtain a good feature ranking that score high features highly predictive of the class.  
\[
	m_{i} = \sum_{y \in Y } \sum _{z \in x^{(i)}} p(z, y)log \Big( \frac{p(z,y)}{p(z)p(y)} \Big),
\]
where $Y$ is the set of class labels, and $p(\cdot,\cdot)$ the joint probability distribution.

A kernel $k$ is then obtained by the matrix product 
 \[
	 k =  (f \cdot m^\top),
\]
where $f$ and $m$ are $n \times 1$ column vectors normalized in the range $0$ to $1$, and $k$ results in a $n \times n $ matrix. 

To boost the performance, we introduce a second feature-evaluation metric based on standard deviation~\cite{Guyon:2002} -- capturing the amount of variation or dispersion of features from average -- as follows:
\[
	\Sigma(i,j) = max\left(\sigma^{(i)},\sigma^{(j)}\right) ,
\]
where $\sigma$ being the standard deviation over the samples of $x$, and $\Sigma$ turns out to be a $n \times n $ matrix with values $\in$ [$0$,$1$].

Finally, the adjacency matrix $A$ of the graph $G$ is given by
\begin{equation}\label{eq:alpha}
	A  = \alpha k + (1-\alpha) \Sigma,
\end{equation}
where $\alpha$ is a loading coefficient $\in [0,1]$. The generic entry $a_{ij}$ accounts for how much discriminative are the feature $i$ and $j$ when they are jointly considered; at the same time, $a_{ij}$ can be considered as a weight of the edge connecting the nodes $i$ and $j$ of a graph, where the $i$-th node models the $i$-th feature distribution (we report the sketch of our method in Algorithm \ref{algorithm}). 

\begin{algorithm}[H]
\caption{Eigenvector Centrality Feature Selection (EC-FS)}
\label{algorithm}
\begin{algorithmic}
\REQUIRE{$X = \{ x^{(1)}, ..., x^{(n)} \}$ , $Y = \{ y^{(1)}, ..., y^{(n)} \}$}\\
\ENSURE{$v_{0}$ ranking scores for each feature }\\
\textbf{- Building the graph} \\
\STATE	$C_1$ positive class, $C_2$ negative class
\FOR{$i = 1 : n$}
\STATE	Compute $\mu_{i,1}$, $\mu_{i,2} $, $\sigma_{i,1} $, and $\sigma_{i,2} $
\STATE	Fisher score: $f(i) = \frac{(\mu_{i,1} - \mu_{i,2})^2}{ \sigma_{i,1}^2 + \sigma_{i,2}^2} $
\STATE	Mutual Information: $m(i) = \sum_{y \in Y } \sum _{z \in x^{(i)}} p(z, y)log \Big( \frac{p(z,y)}{p(z)p(y)} \Big)$
\ENDFOR
\FOR{$i = 1 : n$}
\FOR{$j = 1 : n$}
\STATE  $k(i,j) =  f(i)m(j)$,
\STATE $\Sigma(i,j) = max\left(\sigma^{(i)},\sigma^{(j)}\right)$ ,
\STATE $A(i,j) = \alpha k(i,j) + (1-\alpha) \Sigma(i,j)$
\ENDFOR
\ENDFOR
\STATE \textbf{- Ranking}
\STATE Compute eigenvalues $ \left\lbrace \Lambda\right\rbrace $ and eigenvectors $ \left\lbrace V \right\rbrace $ of $A$\\
\STATE $\lambda_{0} = \underset{\lambda \in \Lambda }{max} (abs(\lambda))$
\RETURN $v_{0}$ the eigenvector associated to $\lambda_{0}$
\end{algorithmic}
\end{algorithm}
\subsection{Eigenvector Centrality}\label{sec:EV}

From a graph theory perspective identifying the most important nodes corresponds to individuate some indicators of centrality within a graph (e.g., the relative importance of nodes). A first way used in graph theory is to study accessibility of nodes, see~\cite{Garrison1960,Pitts1965} for example. The idea is to compute $A^l$ for some suitably large $l$ (often the diameter of the graph), and then use the row sums of its entries as a measure of accessibility (i.e. $scores(i) = [A^l\textbf{e}]_i$, where $\textbf{e}$ is a vector with all entries equal to $1$). The accessibility index of node $i$ would thus be the sum of the entries in the $i$-th row of $A^l$, and this is the total number of paths of length $l$ (allowing stopovers) from node $i$ to all nodes in the graph. One problem with this method is that the integer $l$ seems arbitrary. However, as we count longer and longer paths, this measure of accessibility converges to a index known as eigenvector centrality measure (EC)~\cite{Bonacich}. 

The basic idea behind the EC is to calculate $v_0$ the eigenvector of $A$ associated to the largest eigenvalue. Its values are representative of how strongly each node is connected to the other nodes. Since the limit of $A^l$ as $l$ approaches a large positive number $L$ converges to $v_0$,
\begin{equation}\label{eq:extra}
          \lim_{l \to L} [A^l\textbf{e}] = v_0, 
\end{equation}
the EC index makes the estimation of indicators of centrality free of manual tuning over $l$, and computationally efficient.

Let us consider a vector, for example $\textbf{e}$, that is \emph{not} orthogonal to the principal vector $v_0$ of $A$. It is always possible to decompose $\textbf{e}$ using the eigenvectors as basis with a coefficient $\beta_{0} \neq 0 $ for $v_{0}$.
%
Hence:
\begin{equation}
	\textbf{e} = \beta_0 v_0+\beta_1 v_1+ \ldots +\beta_n v_n,  \quad (\beta_0 \neq 0).
	\label{eq:a1}
\end{equation}
Then
\begin{equation}
\begin{split}
	A \textbf{e} = A(\beta_0 v_0+\beta_1 v_1+ \ldots +\beta_n v_n) = \beta_0 A v_0+\beta_1 A v_1+ \ldots +\beta_n A v_n =	\\
	 = \beta_0 \lambda _0 v_0 +\beta_1 \lambda  _1 v_1+ \ldots +\beta_n \lambda  _n v_n.
\end{split}
	\label{eq:a2}
\end{equation}
So in the same way:
\begin{equation}
\begin{split}
	A^l \textbf{e} = A^l(\beta_0 v_0+\beta_1 v_1+ \ldots +\beta_n v_n) = \beta_0 A^{l} v_0+\beta_1 A^{l} v_1+ \ldots +\beta_n A^{l} v_n =	\\
	 = \beta_0 \lambda _0 ^{l}v_0 +\beta_1 \lambda ^{l}_1 v_1+ \ldots +\beta_n \lambda   ^{l}_n v_n , \quad (\beta_0 \neq 0).
\end{split}
	\label{eq:a3}
\end{equation}
Finally we divide by the constant $\lambda_0^l \neq 0$ (see Perron-Frobenius theorem~\cite{Meyer:2000:MAA:343374}),  
 \begin{equation}
	\frac{A^l \textbf{e}}{\lambda_0^l } = \beta_0 v_0+ \frac{\lambda_1^l \beta_1 v_1}{\lambda_0^l}+ \ldots + \frac{\lambda_n^l \beta_n v_n}{\lambda_0^l}, \quad (\beta_0 \neq 0).
	\label{eq:a4}
\end{equation}
The limit of $\frac{A^l \textbf{e}}{\lambda_0^l }$ as $l$ approaches infinity equals $\beta_0 v_0$ since $ \lim_{l \to \infty} \frac{\lambda_1^l}{\lambda_0^l} = 0 $, $\forall l > 0 $. What we see here is that as we let $l$ increase, the ratio of the components of $A^l\textbf{e}$ converges to $v_0$. Therefore, marginalizing over the columns of $A^l$, with a sufficiently large $l$, corresponds to calculate the principal eigenvector of matrix $A$~\cite{Bonacich}.
\begin{figure}[t!]
\centerline{\includegraphics[width=1\columnwidth]{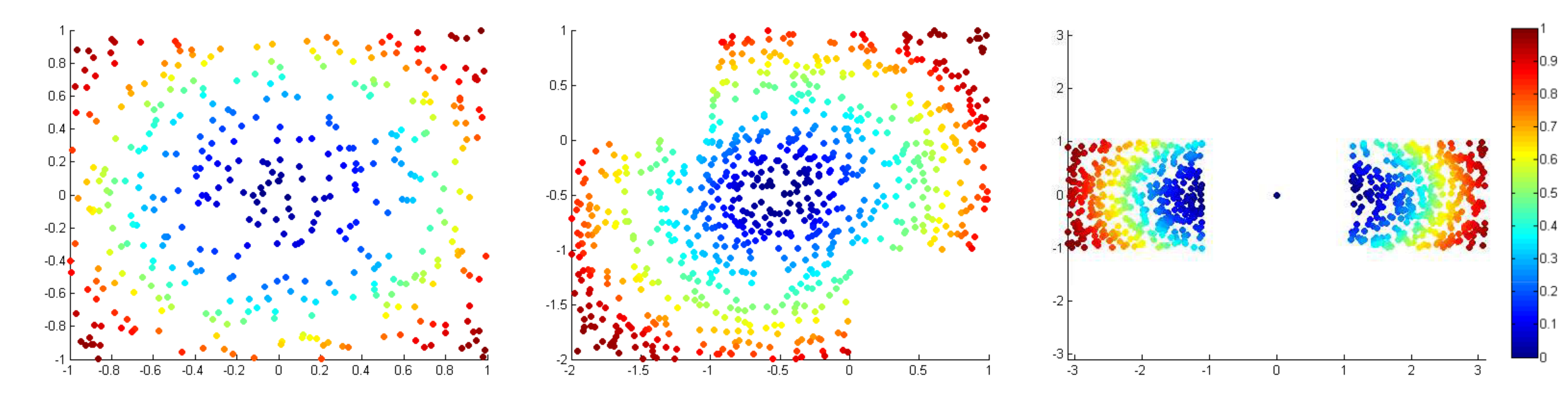}}
\caption{Eigenvector centrality plots for three random planar graphs. On the left, a simple Gaussian distribution where central nodes are at the peripheral part of the distribution as expected. The central and right plots, some more complicated distributions, a node receiving many links does not necessarily have a high eigenvector centrality. Best viewed in color.}
\label{fig:ex}
\end{figure}
Figure~\ref{fig:ex} illustrates a toy example of three random planar graphs. Graphs are made of $700$ nodes and they are weighted by the Euclidean distance between each pair of points. In the example, high scoring nodes are those ones farther from the mean (i.e., the distance is conceived as quantity to maximize), the peculiarity of the eigenvector centrality is that a node is important if it is linked to by other important nodes (higher scores). 

To the aim of this work, the use of eigenvector centrality allows to individuate candidate features, which turn out to be effective from a classification point of view, since indicators of centrality characterize the \emph{global} (as opposed to local) prominence of a feature in the graph. Summarizing, the gist of eigenvector centrality is to compute the centrality of a node as a function of the centralities of its neighbors.

\section{Experiments and Results}\label{sec:exp}\vspace{-0.3cm}

\subsection{Datasets and Comparative Approaches}\vspace{-0.2cm}

The datasets are chosen for letting the proposed method deal with diverse FS scenarios, 
as shown on Table~\ref{table:datasets}. In the details, we consider the problems of dealing with few training samples and many features (\emph{few train} in the table), unbalanced classes (\emph{unbalanced}), or classes that severely overlap (\emph{overlap}), or whose samples are noisy (\emph{noise}) due to: a) complex scenes where the object to be classified is located (as in the VOC series) or b) many outliers (as in the genetic datasets, where samples are often \emph{contaminated}, that is, artefacts are injected into the data during the creation of the samples). Lastly we consider the \emph{shift} problem, where the samples used for the test are not congruent (coming from the same experimental conditions) with the training data.
 
\begin{table*}[t!]
\small
\centering
\begin{tabular}{l c c c c c c c c }
\hline \hline
Name &   \# samples & \# classes & \# feat.  & \emph{few train}& \emph{unbal. (+/-)} & \emph{overlap} & \emph{noise} & \emph{shift} \\\hline
GINA~\cite{GINA} & 3153 &2& 970   &   &  & X & & \\ 
MADELON~\cite{NIPS2003} & 4.4K  &2& 500   &   &  & X &  & \\
\hline
\emph{Colon}~\cite{alon} & 62 &2& 2K   &  X & (40/22) & & X  &  \\
\emph{Lymphoma}~\cite{Golub99} & 45 &2& 4026   &  X & (23/22)& & & \\
\emph{Prostate}~\cite{citeulike:1624492} & 102& 2& 6034   &  X & (50/52)  &  & &\\
\emph{Leukemia}~\cite{Golub99} & 72 &2& 7129   &  X & (47/25) & & X & X \\
\hline \vspace{0.02cm}
VOC 2007~\cite{pascal-voc-2007} & ~10K &20& n.s.   & & X & & X &  \\
\hline
\end{tabular}
\caption{This table reports several attributes of the datasets used.  The abbreviation \emph{n.s.} stands for \emph{not specified} (for example, in the object recognition datasets, the features are not given in advance).}
\label{table:datasets}
\vspace{-5.5mm}
\end{table*}
Table~\ref{table:compmethods} lists the methods in comparison, whose details can be found in Sec.~\ref{sec:SoA}. Here we just note their \emph{type}, that is, \emph{f} = filters, \emph{w} = wrappers, \emph{e} = embedded methods, and their \emph{class}, that is, \emph{s} = supervised or \emph{u} = unsupervised (using or not using the labels associated with the training samples in the ranking operation).  Additionally, we report their computational complexity (if it is documented in the literature). The computational complexity of our approach is $O(Tn +n^2)$.
\begin{table*}[t!]
\small
\centering
\begin{tabular}{p{2.9cm} C{0.7cm} C{0.7cm} C{2.5cm}}
\hline \hline
Acronym &   \small{Type} & \small{Cl.} & Compl.  \\\hline
Fisher~\cite{Quanquanjournals}   &f&s& $\mathcal{O}(Tn)$   \\
FSV~\cite{Bradley98featureselection,Grinblat:2010} & e& s& N/A \\
Inf-FS~\cite{Roffo_2015_ICCV}  &f&u& $\mathcal{O}(n^{2.37}(1+T) )$ \\
MI~\cite{Hutter:02feature} &f& s&$\sim\mathcal{O}( n^2T^2)$ \\
LS~\cite{HCN05a} &f&u& $N/A$   \\
Relief-F~\cite{liu2008} &f&s& $\mathcal{O}(iTnC)$  \\
RFE \cite{Guyon:2002} & w/e& s& \small{$\mathcal{O}(T^2 n log_2n )$}\\
\textbf{Ours}  &f&s& $\mathcal{O}(Tn +n^2)$  \\
\end{tabular}
\caption{List of the FS approaches considered in the experiments, specified according to their \emph{Type}, class (\emph{Cl.}), and complexity (\emph{Compl.}). As for the complexity, $T$ is the number of samples, $n$ is the number of initial features, $K$ is a multiplicative constant, $i$ is the number of iterations in the case of iterative algorithms, and $C$ is the number of classes. N/A indicates that the computational complexity is not specified in the reference paper.}
\label{table:compmethods}
\vspace{-7.5mm}
\end{table*}
The term $Tn$ is due to the computation of the mean values among the $T$ samples of every feature $(n)$. The $n^2$ concerns the construction of the matrix $A$. As for the computation of the leading eigenvector, it costs $O(m^2n)$, where $m$ is a number much smaller than $n$ that is selected within the algorithm~\cite{lehoucq1998arpack}. In the case that the algorithm can not be executed on a single computer, we refer the reader to~\cite{Kang_centralitiesin,Lerman:2010,Rawat2015,MapReduceBetweenness} for distributed algorithms.

\subsection{Exp. 1: Deep Representation (CNN) with pre-training}

This section proposes a set of tests on the PASCAL VOC-2007~\cite{pascal-voc-2007} dataset. 
\begin{table}[t!]
\small
\centering
\begin{tabular}{|C{0.44cm}|C{0.62cm} |C{0.62cm}|C{0.62cm} |C{0.62cm}| C{0.62cm}| C{0.62cm}| C{0.62cm}| C{0.52cm}| C{0.02cm}| C{0.62cm}| C{0.62cm}| C{0.62cm}| C{0.52cm} |C{0.62cm}| C{0.62cm}| C{0.62cm}| C{0.62cm} |  }
\hline
\multicolumn{18}{|c|}{\textbf{PASCAL VOC 2007
}} \\
\hline
 & \multicolumn{8}{c|}{ \textbf{First 128/4096 Features Selected}} & &\multicolumn{8}{c|}{ \textbf{First 256/4096 Features Selected}} \\
\hline
& \tiny{Fisher} &\tiny{FSV} &\tiny{Inf-FS} & \tiny{LS} & \scriptsize{MI} & \tiny{ReliefF} &\tiny{RFE} & \scriptsize{\textbf{Ours}}  & &\tiny{Fisher} &\tiny{FSV} & \tiny{Inf-FS} & \tiny{LS} & \tiny{MI} & \tiny{ReliefF} &\tiny{RFE} & \scriptsize{\textbf{Ours}}  \\
 \hline
\centering
	$ \vcenter{\includegraphics[scale=0.1]{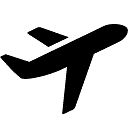}} $ & \scriptsize{52.43} & \scriptsize{	87.90} & \scriptsize{	88.96} & \scriptsize{	\textbf{89.37}} & \scriptsize{	12.84} & \scriptsize{	57.20} & \scriptsize{	86.42} & \scriptsize{	88.09} & 
 & 
\scriptsize{82.65} & \scriptsize{	90.22} & \scriptsize{	\textbf{91.16}} & \scriptsize{	90.94} & \scriptsize{	73.51} & \scriptsize{	81.67} & \scriptsize{	88.17} & \scriptsize{	90.79	}
\\
\hline 
\centering
	$ \vcenter{\includegraphics[scale=0.1]{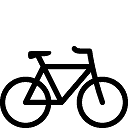}} $ & \scriptsize{ 13.49} & \scriptsize{	80.74} & \scriptsize{	80.43} & \scriptsize{	80.56} & \scriptsize{	13.49} & \scriptsize{	49.10} & \scriptsize{	\textbf{82.14}} & \scriptsize{	80.94} &  & 
\scriptsize{83.21} & \scriptsize{	80.07} & \scriptsize{	83.36} & \scriptsize{	84.21} & \scriptsize{	75.04} & \scriptsize{	71.27} & \scriptsize{	83.30} & \scriptsize{	\textbf{84.72}}
\\
\hline 
\centering
	$ \vcenter{\includegraphics[scale=0.1]{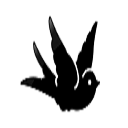}} $ &  \scriptsize{ 85.46} & \scriptsize{	86.77} & \scriptsize{	87.04} & \scriptsize{	86.96} & \scriptsize{	80.91} & \scriptsize{	75.42} & \scriptsize{	83.16} & \scriptsize{	\textbf{88.74}} &  & 
\scriptsize{ 89.14} & \scriptsize{	86.15} & \scriptsize{	88.88} & \scriptsize{	\textbf{89.31}} & \scriptsize{	85.48} & \scriptsize{	83.54} & \scriptsize{	86.12} & \scriptsize{	89.15	}
\\
\hline 
\centering
	$ \vcenter{\includegraphics[scale=0.1]{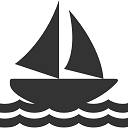}} $ & \scriptsize{ 79.04} & \scriptsize{	83.58} & \scriptsize{	85.31} & \scriptsize{	83.51} & \scriptsize{	61.50} & \scriptsize{	63.75} & \scriptsize{	78.55} & \scriptsize{	\textbf{86.90}} &   & 
\scriptsize{87.05} & \scriptsize{	80.68} & \scriptsize{	87.24} & \scriptsize{	\textbf{87.84}} & \scriptsize{	75.25} & \scriptsize{	73.30} & \scriptsize{	86.13} & \scriptsize{	87.42}
\\
\hline 
\centering
	$ \vcenter{\includegraphics[scale=0.1]{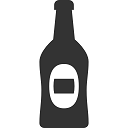}} $ &\scriptsize{ 46.61} & \scriptsize{	39.80} & \scriptsize{	44.83} & \scriptsize{	\textbf{49.36}} & \scriptsize{	35.39} & \scriptsize{	18.33} & \scriptsize{	46.24} & \scriptsize{	47.37} & & 
\scriptsize{52.54} & \scriptsize{	49.00} & \scriptsize{	52.65} & \scriptsize{	49.44} & \scriptsize{	48.94} & \scriptsize{	35.67} & \scriptsize{	47.28} & \scriptsize{	\textbf{53.20}	}
\\
\hline 
\centering
	$ \vcenter{\includegraphics[scale=0.1]{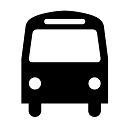}} $ & \scriptsize{ 12.29} & \scriptsize{	72.89} & \scriptsize{	76.69} & \scriptsize{	\textbf{76.98}} & \scriptsize{	12.29} & \scriptsize{	31.54} & \scriptsize{	74.68} & \scriptsize{	76.27} &  & 
\scriptsize{77.32} & \scriptsize{	78.69} & \scriptsize{	79.23} & \scriptsize{	79.97} & \scriptsize{	59.23} & \scriptsize{	63.83} & \scriptsize{	79.38} & \scriptsize{	\textbf{80.57	}}
\\
\hline 
\centering
	$ \vcenter{\includegraphics[scale=0.1]{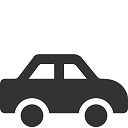}} $ & \scriptsize{ 82.09} & \scriptsize{	78.61} & \scriptsize{	85.78} & \scriptsize{	85.82} & \scriptsize{	63.58} & \scriptsize{	74.95} & \scriptsize{	83.94} & \scriptsize{	\textbf{85.92}} &  & 
\scriptsize{ 85.86} & \scriptsize{	84.01} & \scriptsize{	86.74} & \scriptsize{	\textbf{87.06}} & \scriptsize{	85.27} & \scriptsize{	82.76} & \scriptsize{	85.61} & \scriptsize{	86.56	}
\\
\hline 
\centering
	$ \vcenter{\includegraphics[scale=0.1]{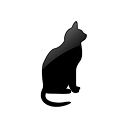}} $ & \scriptsize{ 75.29} & \scriptsize{	82.25} & \scriptsize{	\textbf{83.34}} & \scriptsize{	81.81} & \scriptsize{	40.96} & \scriptsize{	66.95} & \scriptsize{	81.02} & \scriptsize{	83.29} &  & 
\scriptsize{83.46} & \scriptsize{	83.49} & \scriptsize{	\textbf{85.61}} & \scriptsize{	84.98} & \scriptsize{	79.16} & \scriptsize{	76.78} & \scriptsize{	84.50} & \scriptsize{	85.57	}
\\
\hline 
\centering
	$ \vcenter{\includegraphics[scale=0.1]{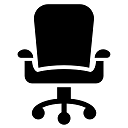}} $  & \scriptsize{ 54.81} & \scriptsize{	52.37} & \scriptsize{	58.62} & \scriptsize{	60.07} & \scriptsize{	16.95} & \scriptsize{	29.07} & \scriptsize{	59.84} & \scriptsize{	\textbf{60.57}} & & 
\scriptsize{63.14} & \scriptsize{	62.54} & \scriptsize{	63.93} & \scriptsize{	64.23} & \scriptsize{	63.20} & \scriptsize{	48.19} & \scriptsize{	62.16} & \scriptsize{	\textbf{64.53}	}
\\
\hline 
\centering
	$ \vcenter{\includegraphics[scale=0.1]{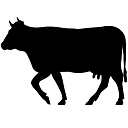}} $ & \scriptsize{ 47.98} & \scriptsize{	61.68} & \scriptsize{	59.23} & \scriptsize{	\textbf{65.50}} & \scriptsize{	11.42} & \scriptsize{	11.42} & \scriptsize{	62.96} & \scriptsize{	60.55} &  & 
\scriptsize{66.51} & \scriptsize{	70.18} & \scriptsize{	67.96} & \scriptsize{	\textbf{71.54}} & \scriptsize{	22.96} & \scriptsize{	51.28} & \scriptsize{	64.20} & \scriptsize{	69.71	}
\\
\hline 
\centering
	$ \vcenter{\includegraphics[scale=0.1]{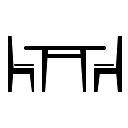}} $ & \scriptsize{49.68} & \scriptsize{	63.50} & \scriptsize{	67.69} & \scriptsize{	63.86} & \scriptsize{	12.62} & \scriptsize{	12.62} & \scriptsize{	67.05} & \scriptsize{	\textbf{67.70}} &   & 
\scriptsize{68.42} & \scriptsize{	69.27} & \scriptsize{	\textbf{71.78}} & \scriptsize{	71.01} & \scriptsize{	65.77} & \scriptsize{	52.24} & \scriptsize{	71.43} & \scriptsize{	70.95	}
\\
\hline 
\centering
	$ \vcenter{\includegraphics[scale=0.1]{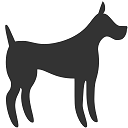}} $ & \scriptsize{ 81.06} & \scriptsize{	80.57} & \scriptsize{	83.16} & \scriptsize{	\textbf{83.21}} & \scriptsize{	70.70} & \scriptsize{	68.12} & \scriptsize{	80.07} & \scriptsize{	83.00} &  & 
\scriptsize{84.24} & \scriptsize{	84.15} & \scriptsize{	85.08} & \scriptsize{	\textbf{85.20}} & \scriptsize{	82.03} & \scriptsize{	74.85} & \scriptsize{	83.52} & \scriptsize{	\textbf{85.20}	}
\\
\hline 
\centering
	$ \vcenter{\includegraphics[scale=0.1]{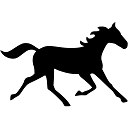}} $& \scriptsize{ 74.91} & \scriptsize{	\textbf{83.33}} & \scriptsize{	81.23} & \scriptsize{	81.75} & \scriptsize{	14.13} & \scriptsize{	63.06} & \scriptsize{	81.55} & \scriptsize{	82.79} & & 
\scriptsize{\textbf{85.68}} & \scriptsize{	83.13} & \scriptsize{	85.28} & \scriptsize{	85.41} & \scriptsize{	71.36} & \scriptsize{	75.53} & \scriptsize{	83.47} & \scriptsize{	85.28	}
\\
\hline 
\centering
	$ \vcenter{\includegraphics[scale=0.1]{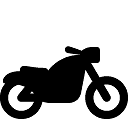}} $ & \scriptsize{ 13.18} & \scriptsize{	71.42} & \scriptsize{	81.32} & \scriptsize{	80.24} & \scriptsize{	13.18} & \scriptsize{	34.43} & \scriptsize{	76.57} & \scriptsize{	\textbf{82.20}} &  & 
\textbf{\scriptsize{84.29}} & \scriptsize{	81.16} & \scriptsize{	84.20} & \scriptsize{	83.81} & \scriptsize{	81.01} & \scriptsize{	70.68} & \scriptsize{	82.97} & \scriptsize{	84.12	}
\\
\hline 
\centering
	$ \vcenter{\includegraphics[scale=0.1]{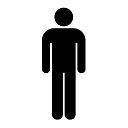}} $ & \scriptsize{ \textbf{91.33}} & \scriptsize{	90.03} & \scriptsize{	89.10} & \scriptsize{	89.33} & \scriptsize{	91.08} & \scriptsize{	88.85} & \scriptsize{	89.03} & \scriptsize{	91.27} &  & 
\scriptsize{91.95} & \scriptsize{	89.99} & \scriptsize{	90.65} & \scriptsize{	90.64} & \scriptsize{	91.77} & \scriptsize{	90.38} & \scriptsize{	90.64} & \scriptsize{	\textbf{91.99}	}
\\
\hline 
\centering
	$ \vcenter{\includegraphics[scale=0.1]{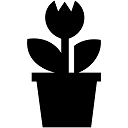}} $ &\scriptsize{ 47.89} & \scriptsize{	39.40} & \scriptsize{	45.38} & \scriptsize{	47.94} & \scriptsize{	13.23} & \scriptsize{	13.30} & \scriptsize{	48.61} & \scriptsize{	\textbf{49.05}} &  & 
\scriptsize{54.94} & \scriptsize{	47.95} & \scriptsize{	53.86} & \scriptsize{	54.31} & \scriptsize{	48.98} & \scriptsize{	34.74} & \scriptsize{	50.18} & \scriptsize{	\textbf{55.88}	}
\\
\hline 
\centering
	$ \vcenter{\includegraphics[scale=0.1]{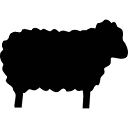}} $ & \scriptsize{ 10.87} & \scriptsize{	68.82} & \scriptsize{	73.35} & \scriptsize{	\textbf{74.05}} & \scriptsize{	10.87} & \scriptsize{	10.87} & \scriptsize{	66.86} & \scriptsize{	73.80} &  & 
\scriptsize{73.43} & \scriptsize{	75.84} & \scriptsize{	79.01} & \scriptsize{	\textbf{81.57}} & \scriptsize{	10.87} & \scriptsize{	11.73} & \scriptsize{	75.47} & \scriptsize{	78.85	}
\\
\hline 
\centering
	$ \vcenter{\includegraphics[scale=0.1]{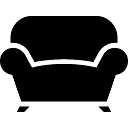}} $ & \scriptsize{ 45.87} & \scriptsize{	56.08} & \scriptsize{	58.94} & \scriptsize{	58.92} & \scriptsize{	13.30} & \scriptsize{	13.31} & \scriptsize{	\textbf{62.06}} & \scriptsize{	61.32} & &
\scriptsize{66.46} & \scriptsize{	59.77} & \scriptsize{	63.07} & \scriptsize{	63.92} & \scriptsize{	58.78} & \scriptsize{	44.74} & \scriptsize{	\textbf{66.68}} & \scriptsize{	64.86	}
\\
\hline 
\centering
	$ \vcenter{\includegraphics[scale=0.1]{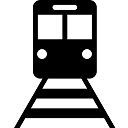}} $ & \scriptsize{ 63.51} & \scriptsize{	88.52} & \scriptsize{	91.42} & \scriptsize{	\textbf{91.48}} & \scriptsize{	58.62} & \scriptsize{	73.32} & \scriptsize{	88.46} & \scriptsize{	91.30} & & 
\scriptsize{84.05} & \scriptsize{	90.61} & \scriptsize{	\textbf{93.21}} & \scriptsize{	93.16} & \scriptsize{	81.33} & \scriptsize{	82.93} & \scriptsize{	90.24} & \scriptsize{	92.31	}
\\
\hline 
\centering
	$ \vcenter{\includegraphics[scale=0.1]{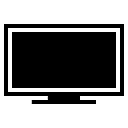}} $  & \scriptsize{ 64.29} & \scriptsize{	65.61} & \scriptsize{	66.79} & \scriptsize{	62.99} & \scriptsize{	47.25} & \scriptsize{	24.96} & \scriptsize{	67.10} & \scriptsize{	\textbf{67.30}} &  & 
\scriptsize{71.44} & \scriptsize{	69.19} & \scriptsize{	70.56} & \scriptsize{	70.75} & \scriptsize{	71.39} & \scriptsize{	55.59} & \scriptsize{	\textbf{73.17}} & \scriptsize{	72.49}
\\
\hline 
\centering
	$ \vcenter{\includegraphics[scale=0.1]{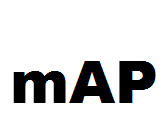}} $  &  \scriptsize{ 54.60} & \scriptsize{	71.69} & \scriptsize{	74.43} & \scriptsize{	74.69} & \scriptsize{	34.72} & \scriptsize{	44.03} & \scriptsize{	73.32} & \scriptsize{	\textbf{75.42}} &  & 
\scriptsize{76.79} & \scriptsize{	75.80} & \scriptsize{	78.17} & \scriptsize{	78.47} & \scriptsize{	66.57} & \scriptsize{	63.09} & \scriptsize{	76.73} & \scriptsize{	\textbf{78.71}	}
\\
\hline 
\end{tabular}
\caption{Varying the cardinality of the selected features. The image classification results achieved in terms of average precision (AP) scores while selecting the first $128$ (3\%) and $256$ (6\%) features from the total $4,096$.}
\label{table:voc07_256}
\vspace{-5.5mm}
\end{table}
In object recognition VOC-2007 is a suitable tool for testing models, therefore, we use it as reference benchmark to assess the strengths and weaknesses of using our approach regarding the classification task. For this reason, we compare our approach against 8 state-of-the-art FS methods reported in Table~\ref{table:compmethods}. 
This experiment considers as features the cues extracted with a deep convolutional neural network architecture (CNN). We selected the pre-trained model called very deep ConvNets~\cite{Simonyan14c}, which performs favorably to the state of the art for classification and detection in the ImageNet Large-Scale Visual Recognition Challenge 2014 (ILSVRC). We use the 4,096-dimension activations of the last layer as image descriptors (i.e., 4,096 features in total). The VOC-2007 edition contains about 10,000 images split into train, validation, and test sets, and labeled with twenty object classes. A one-vs-rest SVM classifier for each class is learnt (where cross-validation is used to find the best parameter C and $\alpha$ mixing coefficient in Eq. \ref{eq:alpha} on the training data) and evaluated independently and the performance is measured as mean Average Precision (mAP) across all classes. 

Table~\ref{table:voc07_256} serves to analyze and empirically clarify how well important features are ranked high by several FS algorithms. The amount of features used for the two experiments is very low: $\approx$3\% and $\approx$6\% of the total. The results are significant: our method achieved the best performance in terms of mean average precision (mAP) followed by the unsupervised filter methods LS and Inf-FS.  
As for the methods in comparison, one can observe the high variability in classification accuracy; indeed, results show that our method is robust to classes (i.e., by changing the testing class its performance is always comparable with the top scoring method).  \vspace{-0.3cm}

\subsection{Exp. 2: Testing on Microarray Databases}

In application fields like biology is inconceivable to devise an analysis procedure which does not comprise a FS step. A clear example can be found in the analysis of expression microarray data, where the expression level of thousands of genes is simultaneously measured. Within this scenario, we tested the proposed approach on four well-known microarray benchmark datasets for two-class problems. Results are reported in Table~\ref{tab_BIO}.  The testing protocol adopted in this experiment consists in splitting the dataset up to 2/3 for training and 1/3 for testing. In order to have a fair evaluation, the feature ranking has been calculated using only the training samples, and then applied to the testing samples. The classification is performed using a linear SVM. For setting the best parameters (C of the linear SVM, and $\alpha$ mixing coefficient) we used a 5-fold cross validation on the training data. This procedure is repeated several times and results are averaged over the trials. Results are reported in terms of the Receiver Operating Characteristic or ROC curves.
\begin{table}[t!]
\begin{center}
\resizebox{1\textwidth}{!}{%
\begin{tabular}{|l|C{0.62cm}|C{0.62cm}|C{0.62cm}|C{0.62cm}|C{0.82cm}|C{0.82cm}|C{0.02cm}|C{0.62cm}|C{0.62cm}|C{0.62cm}|C{0.62cm}|C{0.82cm}|C{0.82cm}|}
\hline
 \multicolumn{14}{|c|}{\large{\textbf{Microarray Databases}}} \\
 \hline
 & \multicolumn{6}{c|}{\textbf{COLON}}                                                                                        & & \multicolumn{6}{c|}{\textbf{LEUKEMIA}}                                                                                         \\ \hline
 & \multicolumn{4}{c|}{\# Features} & \multicolumn{2}{c|}{} & & \multicolumn{4}{c|}{\# Features} & \multicolumn{2}{c|}{} \\ \hline
\multicolumn{1}{|c|}{\scriptsize{\textbf{Method}}}   & 50      & 100     & 150     & 200     & \scriptsize{\textbf{Average}} & \scriptsize{\textbf{Time}} &  & 50      & 100     & 150     & 200     & \scriptsize{\textbf{Average}} & \scriptsize{\textbf{Time}}\\ \hline
\scriptsize{Fisher-S} 
&  \scriptsize{91.25} & \scriptsize{88.44} & \scriptsize{89.38} & \scriptsize{87.81}  & \scriptsize{89.22} & \scriptsize{0.02} & &
\scriptsize{99.33} & \scriptsize{99.78} & \scriptsize{99.78} & \scriptsize{99.78}  & \scriptsize{99.66} & \scriptsize{0.01}  \\ \hline
\scriptsize{FSV} 
&  \scriptsize{85.00} & \scriptsize{88.12} & \scriptsize{89.38} & \scriptsize{89.69}  & \scriptsize{88.04} & \scriptsize{0.18} & &
 \scriptsize{98.22} & \scriptsize{98.44} & \scriptsize{99.11 } & \scriptsize{99.33}  & \scriptsize{98.77} & \scriptsize{0.37}  \\ \hline
\scriptsize{Inf-FS}
&  \scriptsize{88.99 } & \scriptsize{89.41} & \scriptsize{89.32} & \scriptsize{89.01 }  & \scriptsize{89.18} & \scriptsize{0.91} & &
 \scriptsize{99.91} & \scriptsize{\textbf{99.92 }} & \scriptsize{\textbf{99.97}} & \scriptsize{\textbf{99.98}}  & \scriptsize{\textbf{99.95}} & \scriptsize{5.49}  \\ \hline
\scriptsize{LS}
&  \scriptsize{90.31} & \scriptsize{89.06} & \scriptsize{89.38} & \scriptsize{90.00}  & \scriptsize{89.68} & \scriptsize{0.03} & &
\scriptsize{98.67} & \scriptsize{99.33} & \scriptsize{99.56} & \scriptsize{99.56}  & \scriptsize{99.28} & \scriptsize{0.07}  \\ \hline
\scriptsize{MI}
&  \scriptsize{89.38} & \scriptsize{90.31} & \scriptsize{90.63} & \scriptsize{\textbf{90.94}}  & \scriptsize{90.31} & \scriptsize{0.31} & &
 \scriptsize{99.33} & \scriptsize{99.33} & \scriptsize{99.56} & \scriptsize{99.33}  & \scriptsize{98.38} & \scriptsize{0.21}  \\ \hline
\scriptsize{ReliefF}
& \scriptsize{80.94} & \scriptsize{84.38} & \scriptsize{85.94} & \scriptsize{87.50}  & \scriptsize{84.69} & \scriptsize{0.52} & &
 \scriptsize{99.56} & \scriptsize{99.78} & \scriptsize{99.78} & \scriptsize{99.78}  & \scriptsize{99.72} & \scriptsize{1.09}  \\ \hline
\scriptsize{RFE }   
& \scriptsize{89.06} & \scriptsize{85.00} & \scriptsize{86.88} & \scriptsize{85.62}  & \scriptsize{86.64} & \scriptsize{0.18} & &
 \scriptsize{\textbf{100}} & \scriptsize{99.78 } & \scriptsize{99.56} & \scriptsize{99.78}  & \scriptsize{99.78} & \scriptsize{0.14}  \\ \hline
\scriptsize{\textbf{EC-FS} }  
&  \scriptsize{\textbf{91.40}} & \scriptsize{\textbf{91.10}} & \scriptsize{\textbf{91.11}} & \scriptsize{90.63}  & \scriptsize{\textbf{91.06}} & \scriptsize{0.45} & &
 \scriptsize{99.92} & \scriptsize{\textbf{99.92}} & \scriptsize{99.77} & \scriptsize{99.85}  & \scriptsize{99.86} & \scriptsize{1.50}  \\ \hline
\end{tabular}}
\resizebox{1\textwidth}{!}{%
\begin{tabular}{|l|C{0.62cm}|C{0.62cm}|C{0.62cm}|C{0.62cm}|C{0.82cm}|C{0.82cm}|C{0.02cm}|C{0.62cm}|C{0.62cm}|C{0.62cm}|C{0.62cm}|C{0.82cm}|C{0.82cm}|}
\hline
 & \multicolumn{6}{c|}{\textbf{LYMPHOMA}}                                                                                        & & \multicolumn{6}{c|}{\textbf{PROSTATE}}                                                                                         \\ \hline
 & \multicolumn{4}{c|}{\# Features} & \multicolumn{2}{c|}{} & & \multicolumn{4}{c|}{\# Features} & \multicolumn{2}{c|}{} \\ \hline
\multicolumn{1}{|c|}{\scriptsize{\textbf{Method}}}   & 50      & 100     & 150     & 200     & \scriptsize{\textbf{Average}} & \scriptsize{\textbf{Time}} &    & 50      & 100     & 150     & 200     & \scriptsize{\textbf{Average}} & \scriptsize{\textbf{Time}}\\ \hline
\scriptsize{Fisher-S} 
& \scriptsize{98.75} & \scriptsize{98.38} & \scriptsize{98.38} & \scriptsize{100}  & \scriptsize{98.87} & \scriptsize{0.01} & &
 \scriptsize{96.10} & \scriptsize{96.20} & \scriptsize{96.30} & \scriptsize{97.30}  & \scriptsize{96.47} & \scriptsize{0.02}  \\ \hline
\scriptsize{FSV} 
& \scriptsize{98.22} & \scriptsize{98.44} & \scriptsize{99.11} & \scriptsize{99.33}  & \scriptsize{98.77} & \scriptsize{0.18} & &
\scriptsize{96.70} & \scriptsize{96.70} & \scriptsize{96.50} & \scriptsize{96.30}  & \scriptsize{96.55} & \scriptsize{0.63}  \\ \hline
\scriptsize{Inf-FS}
&  \scriptsize{98.12} & \scriptsize{98.75} & \scriptsize{98.75} & \scriptsize{99.38}  & \scriptsize{98.75} & \scriptsize{7.61} & &
 \scriptsize{\textbf{96.80}} & \scriptsize{\textbf{96.90}} & \scriptsize{\textbf{97.10}} & \scriptsize{96.70}  & \scriptsize{96.87} & \scriptsize{26.85}  \\ \hline
\scriptsize{LS}
&\scriptsize{90.00} & \scriptsize{96.88} & \scriptsize{99.38} & \scriptsize{98.75}  & \scriptsize{96.25} & \scriptsize{0.04} & &
 \scriptsize{85.80} & \scriptsize{94.60} & \scriptsize{96.90} & \scriptsize{97.00}  & \scriptsize{93.57} & \scriptsize{0.24}  \\ \hline
\scriptsize{MI}
&  \scriptsize{97.50} & \scriptsize{98.75} & \scriptsize{99.38} & \scriptsize{99.38}  & \scriptsize{98.75} & \scriptsize{0.59} & &
 \scriptsize{96.00} & \scriptsize{\textbf{96.90}} & \scriptsize{96.00} & \scriptsize{96.20}  & \scriptsize{96.27} & \scriptsize{1.01}  \\ \hline
\scriptsize{ReliefF}
&  \scriptsize{96.80} & \scriptsize{97.00} & \scriptsize{98.80} & \scriptsize{98.80}  & \scriptsize{97.85} & \scriptsize{0.74} & &
 \scriptsize{92.72} & \scriptsize{93.46} & \scriptsize{93.62} & \scriptsize{93.85}  & \scriptsize{93.41} & \scriptsize{2.68}  \\ \hline
\scriptsize{RFE }   
&  \scriptsize{96.00} & \scriptsize{98.00} & \scriptsize{98.80} & \scriptsize{99.00}  & \scriptsize{97.95} & \scriptsize{0.02} & &
 \scriptsize{93.40} & \scriptsize{96.40} & \scriptsize{\textbf{97.10}} & \scriptsize{96.32}  & \scriptsize{95.80} & \scriptsize{0.3}  \\ \hline
\scriptsize{\textbf{EC-FS} }  
&  \scriptsize{\textbf{99.40}} & \scriptsize{\textbf{99.20}} & \scriptsize{\textbf{99.60}} & \scriptsize{\textbf{99.20}}  & \scriptsize{\textbf{99.20}} & \scriptsize{1.50} & &
\scriptsize{96.28} & \scriptsize{\textbf{96.90}} & \scriptsize{96.80} & \scriptsize{\textbf{98.10}}  & \scriptsize{\textbf{97.02}} & \scriptsize{2.81}  \\ \hline
\end{tabular}}
\caption{The tables show results obtained on the expression microarray scenario. Tests have been repeated 100 times, and the means of the computed AUCs are reported for each dataset. We indicate with $\epsilon$ each instance where the approach completed the task in less than $0.01$ secs.}
\label{tab_BIO}
\end{center}
\vspace{-12.65mm}
\end{table} 
A widely used measurement that summarizes the ROC curve is the Area Under the ROC Curve (AUC)~\cite{BAMBER1975} which is useful for comparing algorithms independently of application. Hence, classification results for the datasets used show that the proposed approach produces superior results in all the cases. The overall performance indicates that our approach is more robust than the others, by changing the data it still produces high quality rankings. 

\begin{figure*}[!t]
\centerline{\includegraphics[width=0.92\columnwidth]{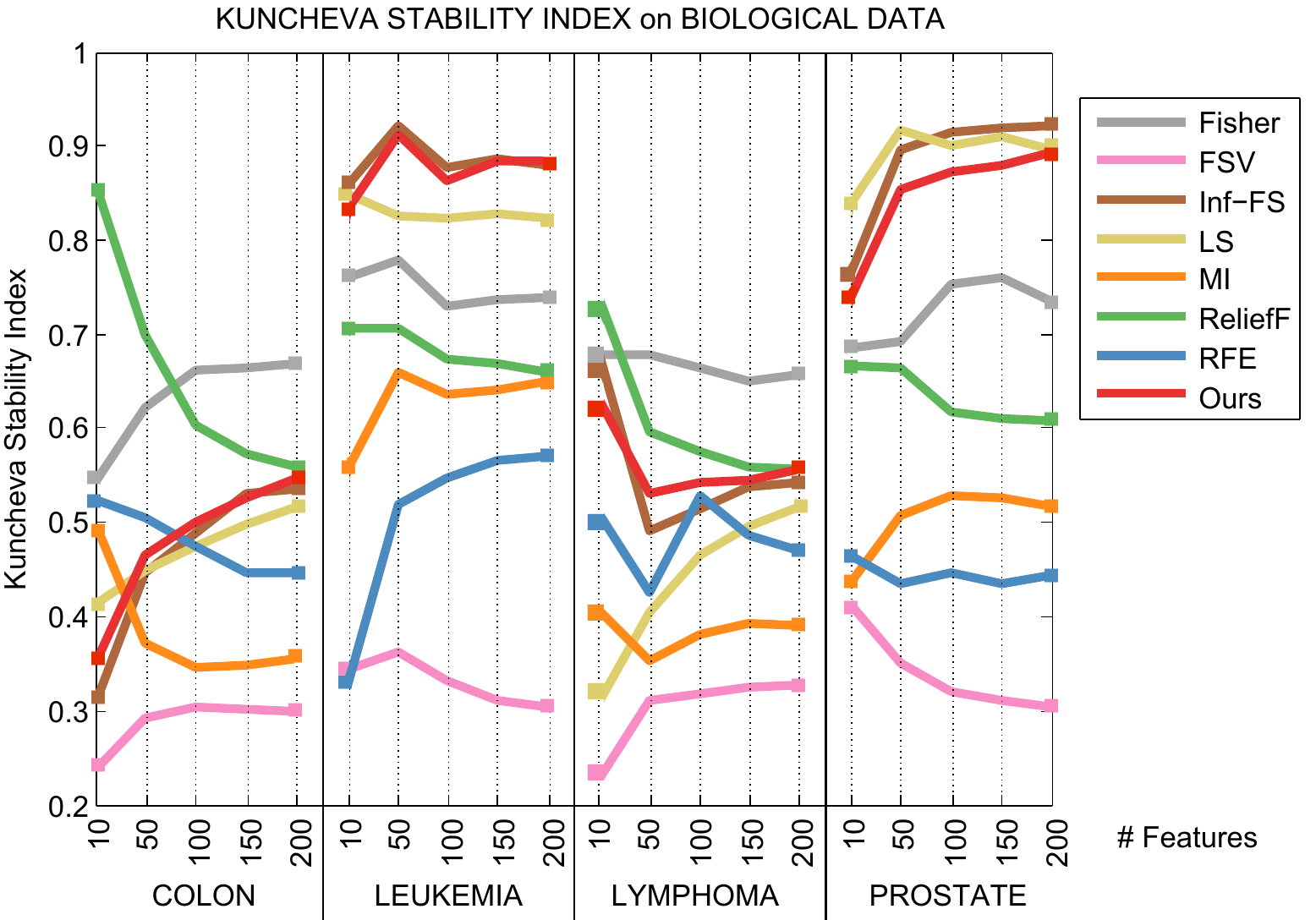}}
\caption{The Kuncheva stability indices for each method in comparison are presented. The figure reports the stability while varying the cardinality of the selected features from 10 to 200 on different benchmarks.}
\label{fig:stability}\vspace{-0.5cm}
\end{figure*}

The quality of a feature subset is measured by an estimate of the classification accuracy of a chosen classifier trained on the candidate subset. Stability of the ranking is an important
aspect when the task is knowledge discovery. The rationale behind this fact is that the estimate of the quality of the candidate subsets usually depends on many the training/testing split of the data. Therefore different sequences of features may be returned from repeated runs of FS approaches. In such a case, it is important to define if these numerous different subsets of features have approximately equal quality, otherwise presenting the user with only one subset may be misleading. We assessed the stability of the selected features using the Kuncheva index~\cite{Kuncheva:2007}. This stability measure represents the similarity between the set of rankings generated over the different splits of the dataset. The similarity between sequences of size $N$ can be seen as the number of elements $n$ they have in common (i.e. the size of their intersection). The Kuncheva index takes values in [-1, 1], and the higher its value, the larger the number of commonly selected features in both sequences. The index is shown in Figure~\ref{fig:stability}, comparing our approach and the other methods. A valid alternative is the stability index based on Jensen-Shannon Divergence $D_{JS}$, proposed by~\cite{guzman2011feature}, with a [0,1] range, where 0 indicates completely random rankings and 1 means stable rankings. Unlike Kuncheva measure, this metric is suitable for different algorithm outcomes: partial sublists (top-k lists) as well as the least studied partial ranked lists. Since in our case we work with full ranked lists, because all the feature selection algorithms considered in this study produce permutations of the original set of features, we preferred the widely used Kuncheva index. 
The proposed method shows, in most of the cases, a high stability whereas the highest performance is achieved.\vspace{-0.3cm}

\subsection{Exp. 2: Other Benchmarks} \vspace{-0.05cm}

GINA has sparse input variables consisting of 970 features. It is a balanced data set
with 49.2\% instances belonging to the positive class. Results obtained on GINA indicate that the proposed approach overcomes the methods in comparison, and select the most useful features from a data set with high-complexity and dimensionality.
\begin{table} 
\begin{center}
\resizebox{1\textwidth}{!}{%
\begin{tabular}{|l|C{0.62cm}|C{0.62cm}|C{0.62cm}|C{0.62cm}|C{0.82cm}|C{0.82cm}|C{0.02cm}|C{0.62cm}|C{0.62cm}|C{0.62cm}|C{0.62cm}|C{0.82cm}|C{0.82cm}|}
\hline
 \multicolumn{14}{|c|}{\large{\textbf{FS Challenge Datasets}}} \\
 \hline
 & \multicolumn{6}{c|}{\textbf{GINA - Handwritten Recognition}} & & \multicolumn{6}{c|}{\textbf{MADELON - Artificial Data}}                                                                                         \\ \hline
 & \multicolumn{4}{c|}{\# Features} & \multicolumn{2}{c|}{} & & \multicolumn{4}{c|}{\# Features} & \multicolumn{2}{c|}{} \\ \hline
\multicolumn{1}{|c|}{\scriptsize{\textbf{Method}}}      & 50      & 100     & 150     & 200     & \scriptsize{\textbf{Average}} & \scriptsize{\textbf{Time}} &  & 50      & 100     & 150     & 200     & \scriptsize{\textbf{Average}} & \scriptsize{\textbf{Time}}\\ \hline
\scriptsize{Fisher-S} 
& \scriptsize{ 89.8} & \scriptsize{ 89.4} & \scriptsize{ 90.2} & \scriptsize{ \textbf{90.4}} & \scriptsize{89.9} & \scriptsize{ 0.05} & &
 \scriptsize{	61.9} & \scriptsize{	63.0} & \scriptsize{62.3} & \scriptsize{64.0} & \scriptsize{   62.5} & \scriptsize{  0.02}  
\\ \hline
\scriptsize{FSV} 
 & \scriptsize{ 81.9} & \scriptsize{ 83.7} & \scriptsize{ 82.0} & \scriptsize{ 83.6} & \scriptsize{ 82.7} & \scriptsize{ 138 } & &
 \scriptsize{59.9} & \scriptsize{ 60.6} & \scriptsize{	61.0} & \scriptsize{ 61.0} & \scriptsize{ 60.7} & \scriptsize{732}  
\\ \hline
\scriptsize{Inf-FS}
 & \scriptsize{ 89.0} & \scriptsize{ 88.7} & \scriptsize{ 89.1} & \scriptsize{ 89.0} & \scriptsize{ 88.9} & \scriptsize{ 41} & &
 \scriptsize{	62.6} & \scriptsize{	\textbf{63.8}} & \scriptsize{	\textbf{65.4}} & \scriptsize{	60.8} & \scriptsize{	63.2} & \scriptsize{  0.04}  \\ \hline
\scriptsize{LS}
& \scriptsize{ 82.2} & \scriptsize{ 82.4} & \scriptsize{ 83.4} & \scriptsize{ 83.2} & \scriptsize{ 82.7} & \scriptsize{ 1.30} & &
 \scriptsize{	62.8} & \scriptsize{	62.9} & \scriptsize{	63.3} & \scriptsize{	64.7} & \scriptsize{	63.4} & \scriptsize{       8.13}  \\ \hline
\scriptsize{MI}
& \scriptsize{ 89.3} & \scriptsize{ 89.7} & \scriptsize{ 89.8} & \scriptsize{ 90.1} & \scriptsize{ 89.6} & \scriptsize{ 1.13} & &
\scriptsize{	63.0} & \scriptsize{	63.7} & \scriptsize{	63.5} & \scriptsize{ 64.7} & \scriptsize{	63.6} & \scriptsize{ 0.4}  \\ \hline
\scriptsize{ReliefF}
 & \scriptsize{ 77.9} & \scriptsize{ 76.3} & \scriptsize{ 77.3} & \scriptsize{ 76.9} & \scriptsize{ 77.2} & \scriptsize{ 0.12} & &
 \scriptsize{ 62.9 } & \scriptsize{ 63.1 } & \scriptsize{	63.2 } & \scriptsize{ \textbf{64.9} } & \scriptsize{ 63.5 } & \scriptsize{ 10.41}
\\ \hline
\scriptsize{RFE }   
 & \scriptsize{ 82.2} & \scriptsize{ 82.4} & \scriptsize{ 83.4} & \scriptsize{ 83.2} & \scriptsize{ 82.7} & \scriptsize{ 6.60} & &
 \scriptsize{  55.0} & \scriptsize{ 61.2} & \scriptsize{ 57.1} & \scriptsize{ 60.2} & \scriptsize{     56.5} & \scriptsize{    50163}  \\ \hline
\scriptsize{\textbf{EC-FS} }  
 & \scriptsize{ \textbf{90.9}} & \scriptsize{ \textbf{90.3}} & \scriptsize{ \textbf{90.4}} & \scriptsize{ 89.5} & \scriptsize{  \textbf{90.3}} & \scriptsize{ 1.56} & &
\scriptsize{	\textbf{63.6}} & \scriptsize{\textbf{63.8}} & \scriptsize{	63.7} & \scriptsize{	63.3} & \scriptsize{	\textbf{63.7}} & \scriptsize{  0.57}  \\ \hline
\end{tabular}}
\caption{Varying the cardinality of the selected features. (ROC) AUC (\%) on different datasets by SVM classification. Performance obtained with the first 50, 100, 150, and 200 features.}
\label{table:NIPS2}
\end{center}
\vspace{-10.5mm}
\end{table}
MADELON is an artificial dataset, which was part of the NIPS $2003$ feature selection challenge. It represents a two-class classification problem with continuous input variables. The difficulty is that the problem is multivariate and highly non-linear. Results are reported in Table~\ref{table:NIPS2}.
This gives a proof about the classification performance of our approach that is attained on the test sets of GINA and MADELON.

FS techniques definitely represent an important class of preprocessing tools, by eliminating uninformative features and strongly reducing the dimension of the problem space, it allows to achieve high performance, useful for practical purposes in those domains where high speed is required.\vspace{-0.3cm}

\section{Reliability and Validity} \vspace{-0.3cm}

In order to assess if the difference in performance is statistically significant, t-tests have been used for comparing the accuracies. Statistical tests are used to determine if the accuracies obtained with the proposed approach are significantly different from the others (whereas both the distribution of values were normal). The test for assessing whether the data come from normal distributions with unknown, but equal, variances is the \emph{Lilliefors} test. Results have been obtained by comparing the results produced by each method over 100 trials (at each trial corresponds a different split of the data). Given the two distributions $x_p$ of the proposed method and $x_c$ of the current competitor, of size $1 \times 100$, a \textit{two-sample t-test} has been applied obtaining a test decision for the null hypothesis that the data in vectors $x_p$ and  $x_c$ comes from independent random samples from normal distributions with equal means and equal but unknown variances. Results (highlighted in Table \ref{tab_BIO} and Table~\ref{table:NIPS2}) show a statistical significant effect in performance (p-value $<$ 0.05, Lilliefors test H=0).\vspace{-0.3cm}

\section{Conclusion}\label{sec:conc}\vspace{-0.3cm}

In this paper we present the idea of solving feature selection via the Eigenvector centrality measure. We design a graph -- where features are the nodes -- weighted by a kernelized adjacency matrix, which draws upon the best-practice in feature selection while assigning scores according to how well features discriminate between classes. The method (supervised) estimates some indicators of centrality identifying the most important features within the graph. The results are remarkable: the proposed method has been extensively tested on $7$ different datasets selected from different scenarios (i.e., object recognition, handwritten recognition, biological data, and synthetic testing datasets), in all the cases we achieve top performances against $7$ competitors selected from recent literature in feature selection.  Our approach is also robust and stable on different splits of the training data, it performs effectively in ranking high the most relevant features, and it has a very competitive complexity. This study also points to many future directions; focusing on the investigation of different implementations for parallel computing for big data analysis or focusing on the investigation of different relations among the features. Finally, we provide an open and portable library of feature selection algorithms, integrating the methods with uniform input and output formats to facilitate large scale performance evaluation. The \textit{Feature Selection Library} (FSLib is available on $Matlab$ $File$ $Exchange$ at \url{https://goo.gl/bvg1ha} ) and interfaces are fully documented. The library integrates directly with MATLAB, a popular language for machine learning and pattern recognition research. \vspace{-0.3cm}

\bibliographystyle{splncs03}
\bibliography{egbib}

\end{document}